# SuperWarp: Supervised Learning and Warping on U-Net for Invariant Subvoxel-Precise Registration

Sean I. Young[✉][iD], Yaël Balbastre[iD], Adrian V. Dalca[iD],
William M. Wells[iD], Juan Eugenio Iglesias[iD], and Bruce Fischl[iD]

Athinoula A. Martinos Center for Biomedical Imaging
and Massachusetts Institute of Technology
`siyoung@mit.edu`   `ybalbastre@mgh.harvard.edu`   `adalca@mit.edu`

**Abstract.** In recent years, learning-based image registration methods have gradually moved away from direct supervision with target warps to self-supervision using segmentations, producing promising results across several benchmarks. In this paper, we argue that the relative failure of supervised registration approaches can in part be blamed on the use of regular U-Nets, which are jointly tasked with feature extraction, feature matching, and estimation of deformation. We introduce one simple but crucial modification to the U-Net that disentangles feature extraction and matching from deformation prediction, allowing the U-Net to warp the features, across levels, as the deformation field is evolved. With this modification, direct supervision using target warps begins to outperform self-supervision approaches that require segmentations, presenting new directions for registration when images do not have segmentations. We hope that our findings in this preliminary workshop paper will re-ignite research interest in supervised image registration techniques. Our code is publicly available from `https://github.com/balbasty/superwarp`.

## 1 Introduction

In recent years, fully convolutional networks (FCNs) have become a universal framework for tackling an array of problems in medical imaging, ranging from image denoising and super-resolution [1, 2] to semantic segmentation [3–5], and from style transfer [6, 7] to image registration. Among these, image registration methods have benefitted immensely from FCNs, allowing methods to transition from an optimization-based paradigm to a learning-based one and to accelerate the alignment of images with different contrasts or modalities, for example.

An overwhelming majority of recent image registration networks [8–11] are trained unsupervised, in the sense that ground-truth deformation fields are not required in the supervision of these networks. Instead, a surrogate photometric loss is used to maximize the similarity between the fixed image and the moving

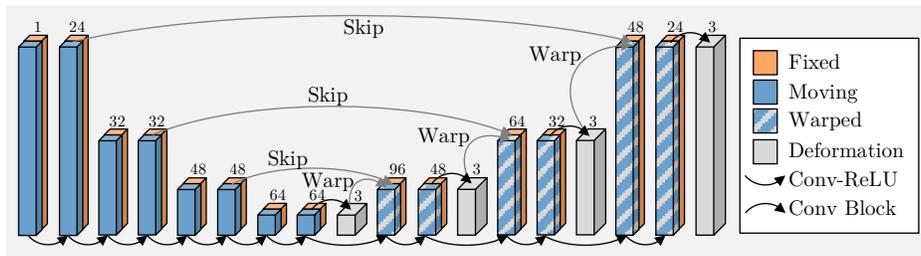

**Fig. 1.** The SuperWarp U-Net for image registration (first four levels shown). The fixed and moving images are concatenated along the batch axis and processed through the network. The two image features are reconcatenated along the channel axis at each level of the U-Net's upward path to be processed into a residual deformation, used to warp the moving features, and scaled and summed to produce the final deformation.

one—warped by the predicted deformation field—in lieu of a loss that penalizes the differences between the predicted and ground-truth deformation fields. Since images typically contain large untextured regions as well as different contrasts and voxel intensities, merely minimizing differences in the fixed image and the moving one is insufficient to recover the ground truth deformation, even when a smoothness prior (or regularization) is imposed on the predicted deformation field. While supervised [12–14] and self-supervised approaches [9, 10]—based on segmentations, for example—produce excellent results, direct supervision using target warps is still desirable in many cases especially if the images do not have segmentations. However, supervised registration has not been as successful for many applications due to severe optimization difficulties faced—the network is jointly tasked with feature extraction and matching in addition to deformation estimation, which is not handled well by a fully convolutional network.

In this work, we will propose SuperWarp, a supervised learning approach to medical image registration. We first re-visit the classic optical flow equation of Horn and Schunck [15] to analyze its implications for supervised registration—the duality of intensity-invariant feature extraction and deformation estimation and the need for multi-scale warping. With such implications in mind, we make one simple but critical modification to the segmentation U-Net that repurposes it for subvoxel- (or subpixel-) accurate supervised image registration. With this modification, direct supervision using target warps outperforms self-supervised registration requiring segmentations. The network, shown in Fig. 1, is strikingly similar to a segmentation U-Net except for warping and deformation extraction layers, allowing U-Net to warp the features as the deformation field is evolved.

## 2   Related Work

SuperWarp is heavily inspired by previous work on optical flow estimation, the aim of which is to recover apparent motion from an image pair [15]. Expressed mathematically, however, optical flow estimation and image registration are in fact identical problems possibly except for the notion of regularity in each—an optical flow field for the former is typically assumed differentiable a.e. whereas a deformation field for the latter infinitely differentiable or diffeomorphic. This subtle distinction between the two problems does however disappear under the supervised learning paradigm since the type of regularity desired is reflected in the ground-truth optical flow (or deformation) fields of the training data.

### 2.1 Optical Flow Estimation

Here, we briefly recap development in classical and learning-based optical flow estimation methods—see e.g. [16] for a review. In their seminal work, Horn and Schunck [15] formulated optical flow estimation via a regularized optimization problem, noting that the problem is generally ill-posed in the absence of local smoothness priors. Several works extend the original Horn–Schunck model [15] using sub-quadratic regularization and data fidelity terms [17–21] that mitigate the deleterious effects of occlusions on flow estimation. Oriented regularization terms [21–25] regularize the flow only along the direction tangent to the image gradient while non-local terms [26–29] regularize flow even across disconnected pixels subject to similar motion. Median filtering of intermediate flows [23, 26] achieves similar effects to non-local regularity terms. Higher-order regularizers [28, 30] assign zero penalty to affine trends in the flow to encourage piecewise-linear flow predictions. Despite the advances, designing a regularizer is highly domain-specific, suggesting that it can be alleviated via supervised learning.

Orthogonally to the choice of regularizers, multi-scale schemes [31–34] have been used to estimate larger flows. Descriptor matching [31, 32] introduces an extra data fidelity term that penalizes misalignment of scale-invariant features (e.g. SIFT), overcoming the deterioration of the conventional data fidelity term at large scales due to the loss of small image structures. Since the optical flow equation no longer holds in the presence of a global brightness change, several authors propose to attenuate the brightness component of the images as a first step using high-pass filters [18, 24, 35], structure-texture decomposition [27, 36] or color space transforms [24]. Thus, in traditional approaches, both multi-scale processing and brightness-invariant transforms require us to handcraft suitable pre-processing filters, which can be highly time-consuming owing to the image-dependent nature of such filters. As we will see, the U-Net architecture used in the SuperWarp obviates the need to handcraft such filters, allowing the U-Net to learn them directly from the training data, end-to-end, to enable brightness-invariant image registration with exceptional generalization ability.

Fischer et al. [37] formulate optical flow estimation as a supervised learning problem. They train a U-Net model to output the optical flow field directly for a pair of input images, supervising the training using the ground-truth optical flow field as the target. Later works extend [37], cascading multiple instances of the network with warping [38], introducing a warping layer [39] or using a fixed image pyramid [40] to improve the accuracy of flow prediction [38, 39] as well as reduce the model size. Some authors propose to tackle optical flow estimation as an unsupervised learning task [41, 42], using a photometric loss to penalize the intensity differences across the fixed and moved images. Recent extensions in this unsupervised direction include occlusion-robust losses [43, 44] based on forward-backward consistency, and self-supervision losses [45, 42]. These are also the building blocks of unsupervised image registration methods [9–11].

## 3 Mathematical Framework

### 3.1 Optical Flow Estimation and Duality Principle

Under a sufficiently high temporal sampling rate, we can relate the intensities of a successive pair of three-dimensional images $(\mathbf{f}_0, \mathbf{f}_1)$ to components $(\mathbf{u}, \mathbf{v}, \mathbf{w})$ of the displacement between the two images using the optical flow equation

$$(\partial \mathbf{f}_1/\partial x) \cdot \mathbf{u} + (\partial \mathbf{f}_1/\partial y) \cdot \mathbf{v} + (\partial \mathbf{f}_1/\partial z) \cdot \mathbf{w} = \mathbf{f}_0 - \mathbf{f}_1 \qquad (1)$$

[15], where $(\partial/\partial x, \partial/\partial y, \partial/\partial z)$ denotes the 3D spatial gradient operator. PDE (1) can also be seen as a linearization of the small deformation model in image registration [46]. Since (1) involves three unknowns for every equation, finding $(\mathbf{u}, \mathbf{v}, \mathbf{w})$ given $(\mathbf{f}_0, \mathbf{f}_1)$ is an ill-posed inverse problem. Smoothness assumptions are therefore made in optimization-based flow estimation [17–21] to render the inverse problem well-posed again similar to image registration [9–11].

A global change in the brightness or contrast across the image pair $(\mathbf{f}_0, \mathbf{f}_1)$ introduces an additive bias in the right-hand side of (1) such that the equation no longer holds. Compensating for this change in pre-processing would require knowledge of the displacement field $(\mathbf{u}, \mathbf{v}, \mathbf{w})$ that we seek in the first place. A similar issue is often met in medical image registration, with different imaging modalities across $\mathbf{f}_0$ and $\mathbf{f}_1$ injecting additive and multiplicative biases in (1). If however we knew the ground-truth displacement $(\mathbf{u}, \mathbf{v}, \mathbf{w})$, harmonizing $(\mathbf{f}_0, \mathbf{f}_1)$ in a normalized intensity space is readily achieved via (1). Conversely, given a harmonized image pair, the displacement field can be recovered using (1).

Image segmentation [47] is the ultimate form of image harmonization, since it removes brightness and contrast from images altogether and turns them into piecewise smooth (constant) signals by construction. This suggests that the use segmentation maps to supervise registration [9, 10] can be beneficial. However, many types of images do not have segmentations available or lack the notion of segmentation altogether, e.g. fMRI activations, so supervision using the ground-truth warps instead can be an expedient way of learning to register.

In practice, images $(\mathbf{f}_0, \mathbf{f}_1)$ are acquired at a low temporal sampling rate so (1) holds only over regions where both image intensities are linear functions of their spatial coordinates [15]. Equivalently, (1) holds in the general case only if the magnitudes of the components $(\mathbf{u}, \mathbf{v}, \mathbf{w})$ are less than one voxel. Since this can pose a major limitation for practical applications, multi-scale processing is used to linearize the images at gradually smaller scales, with the displacement field estimated at the larger scale used to initialize the residual flow estimation at the smaller scale. Linearizing images at larger scales, however, results in the loss of small structures due to the smoothing filters. Handcrafting filters that have an optimum tradeoff between linearization and preservation of image features at every scale is image-dependent and can be time-consuming, implying that learning such filters end-to-end can be beneficial for generalization ability.

### 3.2 Supervised Learning and Multi-scale Warping

SuperWarp exploits the duality principle (1) to supervise an image registration network equipped with multi-scale warping to estimate large deformations. We train a U-Net model on pairs of images with different intensities related via our smoothly synthesized ground-truth deformation fields. The downward path of the U-Net model first extracts intensity-invariant features from the two images separately. The upward path then extracts from the feature pair a deformation that minimizes the differences with respect to the ground-truth target.

SuperWarp makes one important modification to the registration U-Net for large displacement estimation. At each level of the network's upward path, the features of the moving image are first warped using the deformation field from the previous level, such that only the residual deformation, less than a voxel in magnitude, need be extracted at the current level. Processing the two images jointly as a single multi-channel image through U-Net, as done in [9, 10], would entangle the features of the fixed and the moving images, so that warping only the features of the moving one post hoc is not feasible. Instead, we process the two images as a batch with the image pair interacting only during deformation extraction, where the two image features are reconcatenated along the channels axis and processed into deformation field using a convolution block.

Note from the left and the right-hand sides of (1) that it is $(\mathbf{f}_1, \mathbf{f}_0 - \mathbf{f}_1)$, not $(\mathbf{f}_1, \mathbf{f}_0)$, which needs to be processed for displacement estimation. This suggests that feeding the features of $\mathbf{f}_1$ and pre-computed feature differences between $\mathbf{f}_0$ and (warped) $\mathbf{f}_1$ into deformation blocks can yield a saving of one convolution layer per block, which is substantial given that these blocks typically have no more than three convolution layers in total. In practice, we reparameterize the input further to the features of $(\mathbf{f}_0 + \mathbf{f}_1, \mathbf{f}_0 - \mathbf{f}_1)$ to help the extraction blocks average the spatial derivatives of the features across the two images, similarly to the practice in optimization-based approaches [16]. This reparameterization can be seen as a Hadamard transform [48] across the two image feature sets.

### 3.3 Deep Supervision, Data Augmentation and Training

Following the approach of deep supervision for semantic segmentation [49], we supervise the deformation block at each level of the U-Net's upward path with a displacement target. We use the MSE loss between the predicted $(\mathbf{u}, \mathbf{v}, \mathbf{w})$ and the target $(\mathbf{p}, \mathbf{q}, \mathbf{r})$ to minimize $E(\mathbf{u}, \mathbf{v}, \mathbf{w}) = \|(\mathbf{u}, \mathbf{v}, \mathbf{w}) - (\mathbf{p}, \mathbf{q}, \mathbf{r})\|_2^2$. The loss is summed across levels without weighting to produce the final training loss. The deformation block at each level is supervised using the target ground-truth field down-sampled to the spatial dimensions of its predictions. For evaluation, more forgiving EPE loss $E_{\text{EPE}}(\mathbf{u}, \mathbf{v}, \mathbf{w}) = \|(\mathbf{u}, \mathbf{v}, \mathbf{w}) - (\mathbf{p}, \mathbf{q}, \mathbf{r})\|_{2,1}$ is used instead.

**Table 1.** Parameter ranges and probabilities used for random spatial transformation and intensity augmentation of the image pair. Applied separately to each image in the pair.

|  | Spatial Transformation | | | | | Intensity Augmentation | | | |
|---|---|---|---|---|---|---|---|---|---|
|  | Translate | Scale | Rotate | Shear | Elastic | Noise Std | Multiply | Contrast | Gamma |
| **Range** | $\pm 12$ | [0.75,1.25] | $\pm 30°$ | $\pm 0.012$ | $\pm 4$ ($256^2$) | [0,0.05] | [0.75,1.25] | [0.75,1.25] | [0.70,1.50] |
| **Prob.** | 1.0 | 1.0 | 1.0 | 1.0 | 1.0 | 0.5 | 0.5 | 0.5 | 0.5 |

To generate training pairs of images with their corresponding ground-truth deformation targets, we sample an image $\mathbf{f}$ from the training set and synthesize two different smooth displacements $(\mathbf{p}_0, \mathbf{q}_0, \mathbf{r}_0)$ and $(\mathbf{p}_1, \mathbf{q}_1, \mathbf{r}_1)$ that warp $\mathbf{f}$ and produce $\mathbf{f}_0$ and $\mathbf{f}_1$, respectively. The ground-truth displacement is given by

$$(\mathbf{p}, \mathbf{q}, \mathbf{r}) = (\mathbf{Id} + (\mathbf{p}_1, \mathbf{q}_1, \mathbf{r}_1))^{-1}(\mathbf{Id} + (\mathbf{p}_0, \mathbf{q}_0, \mathbf{r}_0)) - \mathbf{Id}, \qquad (2)$$

in which the identity $\mathbf{Id}$ denotes the (vectorization) of the grid coordinates. To facilitate computation, we restrict $(\mathbf{p}_1, \mathbf{q}_1, \mathbf{r}_1)$ to affine fields so that the inverse coordinate mapping $(\cdot)^{-1}$ (2) can be computed by inverting a $4 \times 4$ matrix. We apply a small elastic deformation on $(\mathbf{p}_0, \mathbf{q}_0, \mathbf{r}_0)$ to approximate a higher-order (non-affine) component of the spatial distortion typically seen in MR scans. We then transform the voxel intensities of $\mathbf{f}_0$ and $\mathbf{f}_1$ using a standard augmentation pipeline (Gaussian noise, brightness multiplication, contrast augmentation, and gamma transform); see Table 1 for the hyperparameters of these transforms.

For training, we use a batch size of 1, which actually becomes 2 because the moving and fixed images are concatenated along the batch axis. The Adam [50] optimizer is used with an initial learning rate of $10^{-4}$, linearly reduced to $10^{-6}$ across 200,000 iterations. We find it beneficial to initially train the network for 20,000 iterations on training examples with zero displacement and deformation but still with intensity augmentations to enable the network to learn to extract contrast-invariant features, then introducing deformations to train the network to predict deformations with brightness change across the image pair.

In Fig. 2, we plot validation Dice and end-point error curves of SuperWarp U-Net (ours) and a VoxelMorph-like U-Net baseline for the registration of MR brain scans. In the Dice-supervised case, we train the networks to minimize the regularized Dice loss between the segmentations of the fixed and moving images

$$E_{\text{Dice}}(\mathbf{u}, \mathbf{v}, \mathbf{w}) = D_{\text{Dice}}(\mathbf{f}_1 \circ (\mathbf{Id} + (\mathbf{u}, \mathbf{v}, \mathbf{w})), \mathbf{f}_0) + R(\mathbf{u}, \mathbf{v}, \mathbf{w}), \qquad (3)$$

in which $R$ penalizes the (squared) Laplacian of the components $\mathbf{u}, \mathbf{v}, \mathbf{w}$. In the MSE-supervised case, the same networks are trained to minimize the MSE in the predicted and target deformations. Regardless of the training objective, our SuperWarp U-Net outperforms the baseline U-Net and also trains significantly faster, requiring only 20 iterations to reach maximum accuracy in the case where the Dice loss is used. Moreover, SuperWarp U-Net trained using the MSE loss (no segmentations) outperforms the Dice baseline requiring segmentations.

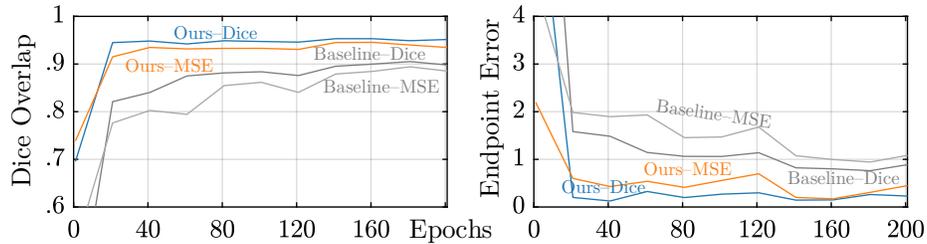

**Fig. 2.** Validation registration accuracy. In both self-supervised (MSE) and supervised (Dice) cases, the SuperWarp U-Net leads to better mean Dice and endpoint error than the baseline (similar to VoxelMorph) and trains faster, requiring only 40 epochs to reach the final accuracy, which are 0.954, 0.152 (Ours–Dice), 0.906, 0.711 (Baseline–Dice).

## 4    Experimental Evaluation

We validate our proposed SuperWarp approaches on two datasets—a set of 2D brain magnetic resonance (MR) scans, as well as the Flying Chairs [37] optical flow dataset widely used in computer vision. The brain image registration task allows us to benchmark the performance of SuperWarp against related work in medical image registration [9, 10] while Flying Chairs allows us to compare the SuperWarp U-Net with the state-of-the-art optical flow estimation networks. In addition to Dice scores between fixed and moved images, we also use the mean EPE to evaluate the accuracy of the displacements. All U-Nets have 7 levels of [24, 32, 48, 64, 96, 128, 192] features and two convolution layers at each level.

### 4.1    Invariant Registration of Brain MR Images

Here, we apply SuperWarp to deformable registration of 2D brain scans within a subject. Obviously, SuperWarp could be applied to the cross-subject setting too but the accuracy of predicted deformations is easier to assess in the within-subject case and facilitates comparisons with other methods. We use the whole brain dataset of [51] containing 40 T1-weighted brain MR scans, along with the corresponding segmentations produced using FreeSurfer [51]. For test, we use a collection of 500 T1-weighted brain MR scans curated from: OASIS, ABIDE-I and -II, ADHD, COBRE, GSP, MCIC, PPMI, and UK Bio. The scan pairs are generated as described in Section 3.3. We do not perform linear registration of the images as a preprocessing step in any of the methods since the displacements are rather small (Table 1) and this provides better insights into their behavior.

To show the improvement in the accuracy of the deformation field recovered using our methods, we plot statistics of the validation end-point error and Dice scores produced by all methods including the baseline—similar to VoxelMorph [9]—in Fig. 3. While our Dice scores are higher than those of the baseline only by 0.04, our end-point errors are more significantly reduced from the baselines (by 80%, on average across, foreground pixels). Fig. 4 shows the displacements predicted by our method, comparing them with those from the baseline.

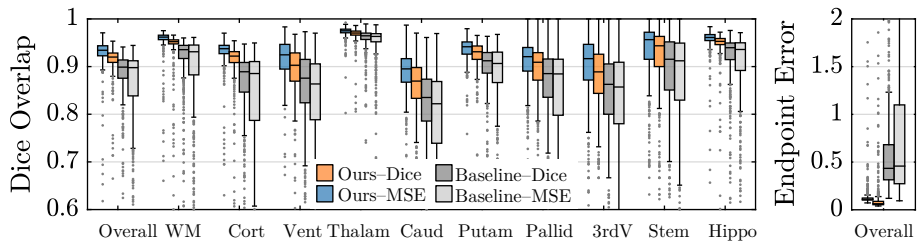

**Fig. 3.** Test Dice (left) and endpoint error (right) statistics on 10 structures across 500 T1w brain images. Regardless of the choice of the training loss function, the SuperWarp produces better Dice and endpoint error than the baseline (similar to VoxelMorph). Note that Ours–MSE does not need or use segmentation information.

To better understand the sources of improvement between the baseline and our approach, we conduct an extensive set of ablation studies on SuperWarp as listed in Table 2. We see that the multi-scale loss used in [49] can actually hurt accuracy for this experiment. Training with the EPE loss produces a worse EPE than training with the MSE loss likely due to numerical instability at zero. The number of U-Net levels should also be high enough (seven) to cover the largest displacements (about $\pm 64$) at the coarsest level of the U-Net.

### 4.2  Optical Flow Estimation

To further benchmark the network architecture used by the SuperWarp, we run additional experiments on the Flying Chairs optical flow dataset [37], popularly used by the computer vision community. To facilitate a fair comparison, we set our network and training hyperparameters very similarly to [39]: 7 U-Net levels for a total of 6.9M learnable parameters, 1M steps, EPE loss for training, multi-scale loss (but weight all scales equally) with the Adam optimizer. Table 3 lists the validation EPE of flow fields predicted using the SuperWarp and other well-performing models.

Both PWC-Net [39] and FlowNet-C [37] attribute their good performance to the use of the cost-volume layer but we find cost volumes to be unnecessary to achieve a good accuracy at least on this dataset. While SPY-Net [40] also uses a multi-scale warping strategy, it is based on a fixed image pyramid. This helps to bring down the number of trainable parameters but can also lead to a loss of image structure at coarser levels. FlowNet2 cascades multiple FlowNet models and warps in between, while the SuperWarp U-Net incorporates warps directly in the model, significantly reducing the model size with a comparable accuracy.

## 5  Discussion

In this paper we have shown that supervising an image registration network with a target warp can achieve state-of-the-art accuracy. Our approach outperforms previous ones due to the multi-scale nature of our prediction—compositions of deformations at each level of the upward path of the U-Net, each applied to the moving image. In this way, each spatial scale receives a moving image as input that has been warped by the composition of all larger spatial scales, ensuring that optic flow condition holds for the deformation at that level. We show that this recovers the accuracy of the deformation estimation that was lost in previous supervised techniques.

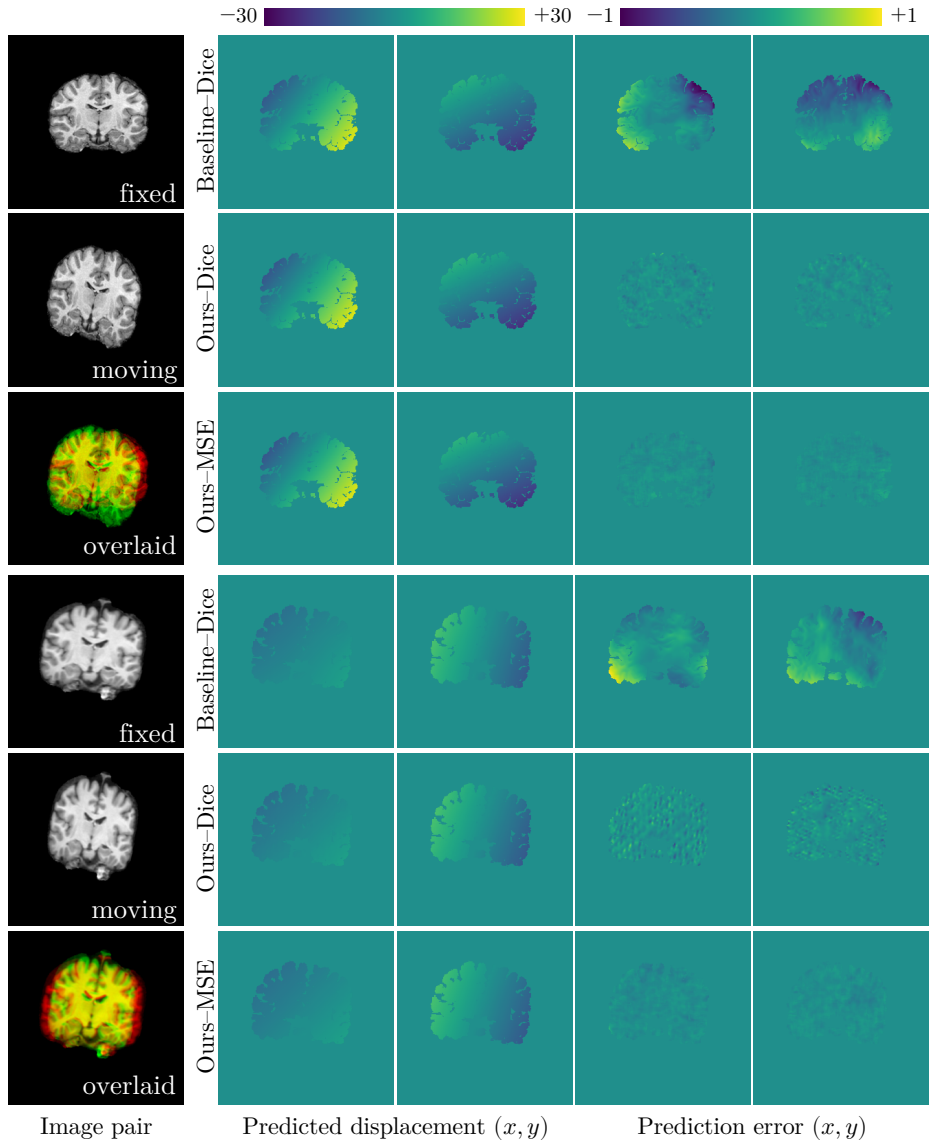

**Fig. 4.** Visualization of predicted displacement fields (test). Both Dice and MSE variants of the SuperWarp can produce highly accurate displacements (in the first example, 0.08 and 0.06 mm, respectively) whereas the baseline (similar to VoxelMorph) prediction has larger errors (0.41 mm). Images are 2D, 256 × 256, 1 mm isotropic. Cf. Fig. 3 (right).

**Table 2.** Ablation of network and training hyperparameters used and their influence on the best epoch validation accuracy. Default hyperparameter: (7, MSE, False, True).

|      | Number of levels | | Training loss function | | | Multi-scale loss | | Multi-scale warp | |
|------|-------|-------|-------|-------|-------|-------|-------|-------|-------|
|      | 6     | 7     | Dice  | EPE   | MSE   | True  | False | True  | False |
| Dice | 0.927 | **0.947** | **0.954** | 0.942 | 0.947 | 0.939 | **0.947** | **0.947** | 0.903 |
| EPE  | 0.450 | **0.122** | **0.103** | 0.195 | 0.122 | 0.270 | **0.122** | **0.122** | 0.738 |

**Table 3.** Mean EPE achieved by various network models on the Flying Chairs test set.

|            | PWC-Net | SPY-Net | FlowNetS | FlowNetC | FlowNet2 | Ours–EPE |
|------------|---------|---------|----------|----------|----------|----------|
| Parameters | 8.75M   | **1.20M** | 32.1M  | 32.6M    | 64.2M    | 6.9M     |
| EPE        | 2.00    | 2.63    | 2.71     | 2.19     | **1.78** | 1.82     |

While segmentation accuracy is itself of course important, we also point out that there are instances in which it is important to recover an exact deformation field. In these causes, using photometric losses will lead to inaccuracies in regions in which there is not sufficient image texture to guide the estimation. We show that using the architecture we have described, we are able to recover an excellent prediction of a true underlying deformation field. Uses cases include distortion estimation and removal in MRI, such as those caused by inhomogeneities in the main magnetic field (B0) and image distortions induced by nonlinearities in the gradient coils used to encode spatial location.

### 5.1   Future Work

In this workshop paper, we have addressed only one type of invariance, namely invariance to intensity (or illumination) change across images. In the sequel, we plan to add contrast and distortion invariance to the network by training it on synthetic scans of various contrasts as done in [52] and applying the synthetic approach to distortions as well. Also, we plan to run a more comprehensive set of experiments on 3D MR images, showing the benefits of our approach in many clinical applications.

## Acknowledgments

Support for this research provided in part by the BRAIN Initiative Cell Census Network grant U01MH117023, NIBIB (P41EB015896, 1R01EB023281, R01EB-006758, R21EB018907, R01EB019956, P41EB030006, P41EB028741), NIA (1R-56AG064027, 1R01AG064027, 5R01AG008122, R01AG016495, 1R01AG070988), NIMH (R01MH123195, R01MH121885, 1RF1MH123195), NINDS (R01NS05-25851, R21-NS072652, R01NS070963, R01NS083534, 5U01NS086625, 5U24NS-10059103, R01NS105820), ARUK (IRG2019A-003), and was made possible by resources from Shared Instrumentation Grants 1S10RR023401, 1S10RR019307,

and 1S10-RR023043. Additional support was provided by the NIH Blueprint for Neuroscience Research (5U01-MH093765), part of the multi-institutional Human Connectome Project.